\documentclass[pdflatex,sn-mathphys-num]{sn-jnl}

\usepackage{graphicx}
\usepackage{multirow}
\usepackage{amsmath,amssymb,amsfonts}
\usepackage{amsthm}
\usepackage{mathrsfs}
\usepackage[title]{appendix}
\usepackage{xcolor}
\usepackage{textcomp}
\usepackage{manyfoot}
\usepackage{booktabs}
\usepackage{algorithm}
\usepackage{algorithmicx}
\usepackage{algpseudocode}
\usepackage{listings}

\theoremstyle{thmstyleone}

\theoremstyle{thmstyletwo}

\theoremstyle{thmstylethree}

\begin{document}
\title[Article Title]{High-Quality Facial Albedo Generation for 3D Face Reconstruction from a Single Image using a Coarse-to-Fine Approach}
\author[1]{\fnm{Jiashu} \sur{Dai}}\email{daijiashudai@ahpu.edu.cn}
\equalcont{These authors contributed equally to this work.}

\author*[1]{\fnm{Along} \sur{Wang}}\email{wangalong\_jsj@163.com}
\equalcont{These authors contributed equally to this work.}

\author[1]{\fnm{Binfan} \sur{Ni}}\email{binfanni@gmail.com}

\author[1]{\fnm{Tao} \sur{Cao}}\email{taolaw@foxmail.com}

\affil[1]{\orgdiv{School of Computer and Information}, \orgname{Anhui Polytechnic University}, \orgaddress{\street{Guandou}, \city{Wuhu}, \postcode{24100}, \state{Anhui}, \country{China}}}

\abstract{
Facial texture generation is crucial for high-fidelity 3D face reconstruction from a single image. However, existing methods struggle to generate UV albedo maps with high-frequency details.
To address this challenge, we propose a novel end-to-end coarse-to-fine approach for UV albedo map generation. Our method first utilizes a UV Albedo Parametric Model (UVAPM), driven by low-dimensional coefficients, to generate coarse albedo maps with skin tones and low-frequency texture details. To capture high-frequency details, we train a detail generator using a decoupled albedo map dataset, producing high-resolution albedo maps. Extensive experiments demonstrate that our method can generate high-fidelity textures from a single image, outperforming existing methods in terms of texture quality and realism. The code and pre-trained model are publicly available at https://github.com/MVIC-DAI/UVAPM, facilitating reproducibility and further research.
}


\maketitle

\begin{figure}[H]
\centering
\includegraphics[width=1.0\textwidth]{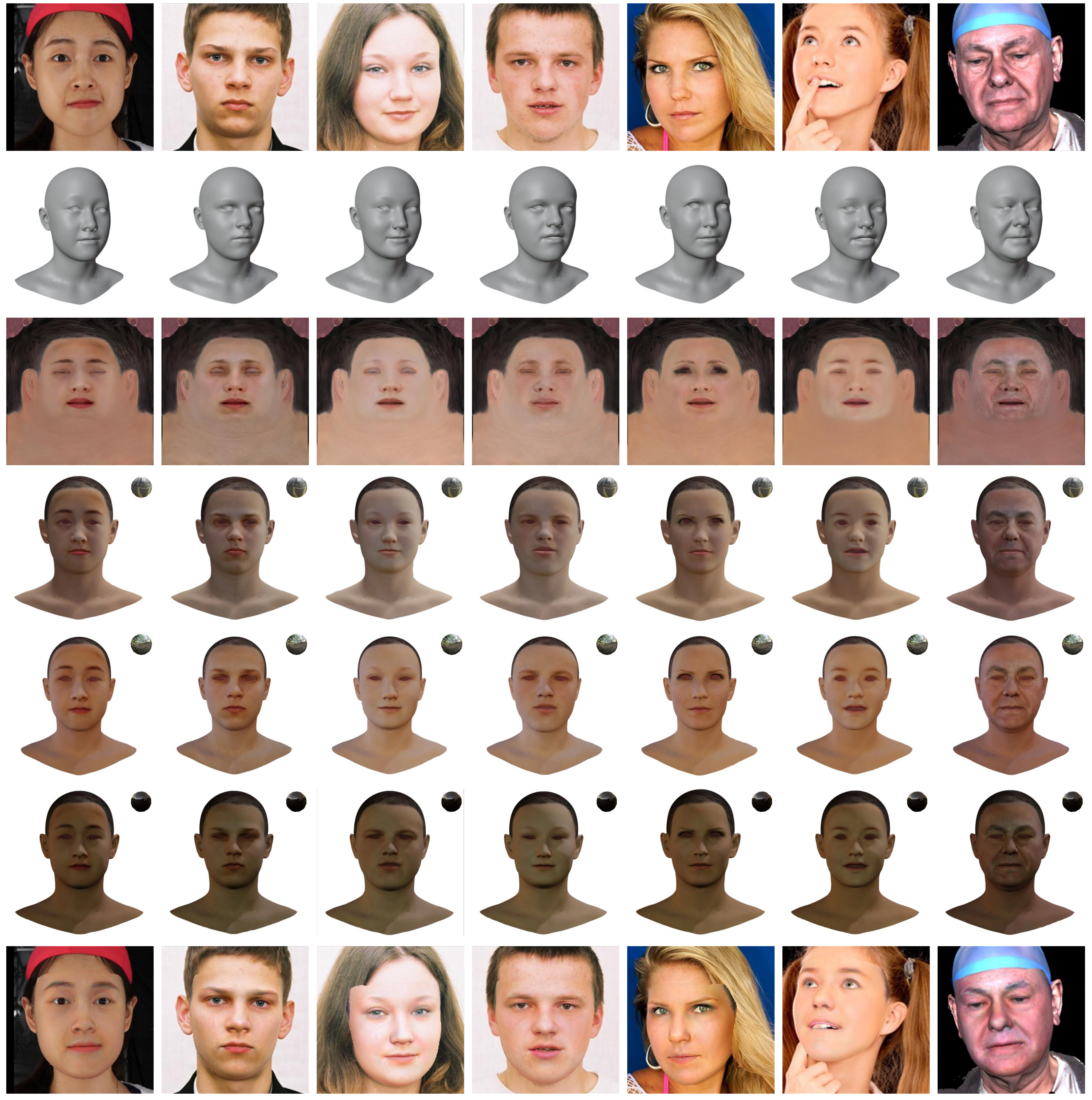}
\caption{Examples of texture images generated using our method. Each set of images in order from top to bottom: real face images; reconstructed shapes; generated UV albedo maps; rendering results in different environments; and rendering results of generated textures.}\label{fig:preview}
\end{figure}
\section{Introduction}

In recent years, 3D face models have played a critical role in VR/AR, face recognition\cite{he2023occlusion}, face parsing\cite{yang20253d}, etc. High-fidelity 3D face reconstruction from a single image has always been an important research direction in computer vision and image processing. 

Shape and texture are the two main components of a 3D face model. Over the past two decades, significant advancements have been made in 3D face reconstruction methods\cite{3DMM, AlbedoGAN, DECA, HRN, Hiface} utilizing the 3D Morphable Model (3DMM), initially introduced by Blanz and Vetter\cite{3DMM}. Their approach demonstrated the feasibility of reconstructing the shape and albedo by fitting the coefficient of the linear statistical model. This innovation effectively transforms the complex task of face reconstruction into a low-dimensional coefficient regression problem.
To improve the shape refinement of the face model, many methods\cite{BFM2009, LYHM, FaceScape, FLAME, HIFI3D} try to increase the densities of vertices and triangles of the face model. These can be roughly divided into two categories. Ones make 3DMM by capturing high-precision face-scanned data or producing a multilinear face model\cite{FaceScape}, and the others\cite{displacement-map, DECA, lee2022holistic, Hiface, HRN} adopt UV displacement maps by nonlinear regression to refine the reconstructed coarse shape.
Although many works have improved the accuracy of shape estimation, only a few have addressed the issue of texture UV map generation. Realistic facial texture can greatly improve the perception of the face model. Several methods\cite{smith2020morphable, HIFI3D} expand the linear solution space of face texture by increasing the diversity of the Principal Component Analysis (PCA)\cite{PCA} texture bases. These methods are guided by prior knowledge that can generate complete textures even under large occlusions. However, since the expressive power of linear models and the density of the vertices in topological models are limited, this results in textures that often lack high-frequency details.

In recent years, some approaches have tried using high-resolution UV albedo maps to improve texture quality. This helps overcome the limitations of vertex density by transforming the task of predicting 3D vertex colors into a Pix2Pix task. 
The AvatarMe\cite{AvatarMe} combines linear texture basis fitting with a super-resolution neural network to create high-resolution maps, enhancing facial texture generation. However, fine details like skin texture and small wrinkles are often not fully captured.
Non-linear generative models have recently achieved great success in synthesizing high-quality face images. Methods like UV-GAN\cite{UV-GAN}, GANfit\cite{Ganfit}, OSTeC\cite{Ostec}, and AlbedoGAN\cite{AlbedoGAN} use GANs to model the distribution of high-resolution UV maps, replacing the linear texture bases in 3DMM. PR3D\cite{huang2024PR3D} proposed a texture extraction method that extracts texture from input face images and images reenacted using StyleGAN2, thereby constructing a facial texture map. Although these methods generate high-resolution texture maps, the facial images in the training datasets often have complex lighting effects. This results in texture maps with unrealistic shadows, which makes them unsuitable for rendering under different lighting conditions.
Nextface\cite{NextFace} uses a multi-stage strategy to predict facial specular reflectance, diffuse reflectance, and Lambertian reflectance. Relightify\cite{Relightify} and UV-IDM\cite{UV-IDM} improve realism and efficiency by using diffusion models for image transformation. However, the low-resolution albedo maps struggle to capture fine details. The performance of the texture decoder relies on the variety, amount, and quality of UV texture data. FaceScape\cite{FaceScape} offers a high-quality, publicly available texture map dataset collected in a controlled environment. However, it is limited to only 938 unique identities. FFHQ-UV\cite{FFHQ-UV} creates a publicly available high-quality UV texture data set that contains over 50,000 UV maps through UV texture extraction, correction, and complementation procedures, using HiFi3D++\cite{REALY} topological model and StyleFlow\cite{styleflow}. The above works advance the facial texture generation task. 

In this paper, we propose a novel end-to-end coarse-to-fine UV albedo map generation method. we introduce a UV Albedo Parametric Model(UVAPM) that can be driven by low-dimensional coefficients for generating coarse albedo maps with a resolution of $256\times256$.
Specifically, we first collect about 100,000 uniformly illuminated, unobstructed, high-resolution ($1024\times1024$) UV albedo maps from FFHQ-UV datasets and produced by its pipeline.
Secondly, we unfold the RGB channels of each UV map separately according to the row-first rule and compute the PCA bases of the channel, rather than mixing all the channels. Finally, we choose the 100 top feature vectors of the covariance matrix in each channel, a total of 300 dimensions. 
Limited by the expressive power of the linear model, the regressed textures tend to be smooth and lack high-frequency texture details such as eye bags, wrinkles, and spots. To be more refined for coarse albedo maps, we introduce a detail generator trained using the Variational Auto-Encoder(VAE)\cite{VAE} and can generate facial details with a resolution of $512\times512$. The reconstructed shapes, the final generated albedo maps, the effects of rendering in different environments, and the rendering back to the input image effects by our method are shown in Figure \ref{fig:preview}.

In summary, our main contributions can be summarized as follows :
\begin{enumerate}
    \item We propose a novel end-to-end high-quality facial albedo generation method that can be used on HiFi3Ds-based face reconstruction methods. 
    \item The UV Albedo Parametric Model(UVAPM) is proposed to generate coarse facial albedo maps that can be driven by low-dimensional coefficients.
    \item A novel detail generator is used to generate high-frequency facial details (e.g., whiskers, wrinkles, spots, etc.).
    \item We conducted extensive experiments on $3$ face image datasets. Our method is compared qualitatively and quantitatively with existing reconstruction methods. The results show that our method can generate texture maps with high-frequency details from a single image with high performance.
\end{enumerate}

\section{Related Work}
\subsection{3D morphable model}

3DMMs\cite{3DMMsynthesis} are statistical models, first introduced by Blanz and Vetter\cite{3DMM}, which are still widely used to constrain the components of 3D faces in 3D face reconstruction tasks. The models are produced from topologically aligned 3D face models by PCA.  Over the past 20 years, various expression and texture models have been proposed to enhance the expressive power of face models.
Paysan et al.\cite{BFM2009} proposed the Basel Face Model (BFM) using the Optical Flow algorithm from 200 scans. gerig et al.\cite{Facewarehouse} added an expression basis to the BFM model and constructed the FaceWarehouse model. In order to expand the linear solution space, booth et al.\cite{LSFM} constructed the LSFM model by collecting 9,663 sets of 3D face data and using the NICP template morphing method. 
High-quality texture in 3D models is crucial. Dai et al.\cite{dai20173d} collected 1,212 individual texture maps on the LYHM\cite{LYHM} dataset employing a pixel embedding method to construct a texture basis, thus replacing vertex-by-vertex color mapping. \cite{smith2020morphable} proposed a light-stage capture and processing pipeline and captured 73 scans. The HiFi3D++\cite{REALY} model was produced from about 2,000 sets of topologically consistent shapes, which further enhances the expressiveness of 3D face models.
FFHQ-UV\cite{FFHQ-UV} using the HiFi3D++ topological model and the StyleGAN-based image editing method produced a public large-scale UV texture dataset that contains over 50,000 high-quality UV texture maps with even illuminations, neutral expressions, high-frequency details, and cleaned facial regions. Provides a reliable source of knowledge for our UVAPM.

In this paper, we use HiFi3D++ as a shape and expression model that can be driven by low-dimensional identity and expression coefficients to obtain a full-head model. In addition, our motivation for proposing UVAPM is inspired by 3DMMs, which are used to generate coarse UV albedo maps that guarantee the accuracy of basic facial skin tones and individual facial feature colors. See Section \ref{sec:UV Albedo Parametric Model} for more details.

\subsection{3D face reconstruction}

Traditional methods using Analysis-by-Synthesis obtain 3DMM coefficients by iterative optimization, which is susceptible to initialization coefficients. With the development of deep learning, many methods\cite{emoca, Deep3dface, SADRNet, shang2023jr2net, lin2020towards, deng2023fast} build encoder-decoder network structures. Among them, the decoder is usually a variety of 3DMMs, and the encoder generally extracts image features using Convolutional Neural Networks(CNNs) and Graph Convolutional Networks(GCNs) to obtain 3DMM, pose, and illumination coefficients, etc. Some methods\cite{FFHQ-UV, HIFI3D, Ganfit, dense-landmarks, FaceScape} optimize the 3DMM coefficients with the help of an optimizer. 
To achieve end-to-end training of the network, differentiable renderers\cite{dib2021towards, differentiable_renderer, zhu2020reda} are introduced for rendering the reconstructed model to the plane, constraining the network by calculating the difference between the input image and the rendered image through various loss functions\cite{wen2021self, emoca, 3DDFA_V3, DECA, dense-landmarks} such as landmark loss, photometric loss, identity loss, etc., and optimizing the network parameters in a self-supervised, weakly-supervised manner. Such methods usually demonstrate excellent generalization ability in complex environments.

Constrained by the linear representation capability of 3DMM, the reconstructed facial model is relatively coarse and smooth, and lacks details such as wrinkles, dimples, etc. Shape-based shading (SfS)\cite{jiang20183d, li2018feature} methods can reconstruct facial details from single-view or multi-view images, however, these methods are sensitive to the self-occlusion, large-angle, and low-light interference. Recent works\cite{displacement-map, ling2022structure, HRN, Hiface, dib2024mosar, liu2023transformer} re-topology the shape of coarse models by predicting UV displacement maps and bringing promising results. Inspired by this idea, we present an end-to-end, coarse-to-fine, multi-stage parameter-optimized texture generation method.

\subsection{Faciel texture generation}

Traditional 3DMM\cite{BFM2009, LSFM} texture bases record the RGB values of per-vertex in the uniform topological model, so the fineness of the facial texture depends on the density of the vertices. At lower densities, 2D pixels belonging to non-vertices will be obtained using interpolation strategies during rasterization, thus making it difficult to show high-frequency texture details in the end.  

In recent years, some methods\cite{Ganfit, UV-GAN, FFHQ-UV} adopt the UV maps as texture representation information in 3D models. UV Mapping is a technique in 3D modeling for unwrapping the surface of a 3D model onto a 2D plane. In this way, the 2D image can be accurately attached to the surface of the 3D model to achieve realistic texture effects. UV Unwrapping is the process of creating UV mappings, such as Planar unwrapping, Cylindrical unwrapping, Spherical unwrapping, etc.  Following a uniform UV mapping function, each vertex of the model is assigned a 2D coordinate in the UV map.
Therefore, the UV map becomes more expressive in texture representation as the resolution increases. In this paper, we generate high-fidelity UV albedo maps using a coarse-to-fine approach. 
\section{Methodology}
\subsection{Preliminary}

\subsubsection{3D face morphable model}
HiFi3D++ is a 3DMM that can be driven by identity$\beta\in\mathbb{R}^{|\beta|} $ and expression$\xi\in\mathbb{R}^{|\xi|}$ coefficients to get a full-head mesh with $n=20,481$ vertices and $f=40,832$ triangles. The model $S$ is: 
\begin{equation}
S = S(\beta,\xi)=\bar{S}+\beta \mathbf{B}_{id}+\xi \mathbf{B}_{exp},
\end{equation}
where $\Bar{S}$ is the mean shape computed by the PCA process. $\mathbf{B}_{id}$ and $\mathbf{B}_{exp}$ are the PCA bases of face identity and expression, respectively. See \cite{HIFI3D, REALY} for details.
\subsubsection{Camera model}
The weak perspective projection model is employed to project 3D vertices onto a 2D plane: 
$$
p=(f\times S_{i}\times \Lambda + t) \times \mathbf{\Pi},
$$
where $p\in\mathbb{R}^2$ denotes a projected 2D coordinate, $S_i\in\mathbb{R}^3$ is a vertex in mesh $S$, $f\in\mathbb{R}^1$ and $t\in\mathbb{R}^3$ denote the scaling factor and translation consists of \{$t_x, t_y, t_z$\}, respectively. $\mathbf{\Pi}\in\mathbb{R}^{3\times2}$ is the orthographic projection matrix. $\Lambda$ is the 3D rotation matrix computed by Euler angles \{$r_{pitch}, r_{yaw}, r_{roll}$\}. The pose coefficient $\theta=\{f, t_x, t_y, t_z, r_{pitch}, r_{yaw}, r_{roll}\}$.

\subsubsection{Illumination Model}
We follow previous work \cite{Hiface, Deep3dface} to use Spherical Harmonics (SH) \cite{SH_model} as our illumination model instead of the Phong reflection model. The shaded texture $T$ is computed as:
$$
T(\alpha, \gamma, N_{uv})_{i,j}=A(\alpha)_{i,j} \odot \sum_{k=1}^{9}\gamma_k\mathbf{H}_k(N_{i,j}),
$$
where the albedo, $A$, surface normals, $N$, and shaded texture, $T$, are represented in UV coordinates and where $T_{i,j}\in\mathbb{R}^3$, $A_{i,j}\in\mathbb{R}^3$, and $N_{i,j}\in\mathbb{R}^3$ denote pixel $(i,j)$ in the UV coordinate system. 
The SH basis and coefficients are defined as $\mathbf{H}_k$ : $\mathbb{R}{^3}\to\mathbb{R}$ and $\gamma\in\mathbb{R}^{3\times9}$ , with $\gamma_k\in\mathbb{R}^3$, and $\odot$ denotes the Hadamard product.

\subsubsection{Rendering}
Given the geometry coefficients\{$\beta, \xi, \theta$\}, albedo $\alpha$ and lighting $\gamma$, we can generate the 2D image $I_r$ by rendering as $I_r=\mathcal{R}(S,T,\theta)$, where $\mathcal{R}$ denotes the differentiable rendering function. HiFi3D++ is able to generate face geometry with various poses, shapes, and expressions from a low-dimensional latent space. 

\subsection{Overview}
\begin{figure}[h]
\centering
\includegraphics[width=1.0\textwidth]{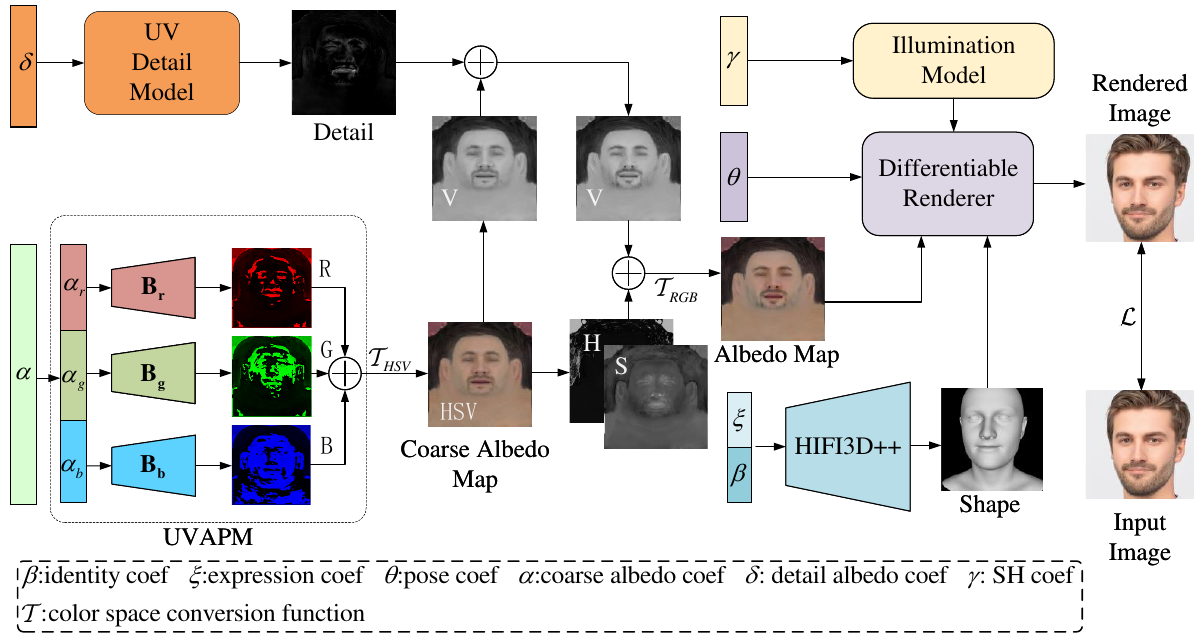}
\caption{Illustration of our coefficient optimization framework. }\label{fig:overview}
\end{figure}

We reconstruct shapes and generate realistic textures employing multistage coefficient optimization.

In stage 1, the identity coefficients, illumination coefficients, and head pose coefficients are regressed by using checkpoints trained by the Deep3D weakly supervised framework. A multilevel loss function is used to further constrain the shape; see \ref{sec:loss_functions} for more details.

In stage 2, a coarse albedo map is generated. UVAPM is driven by $\alpha_c$ to generate a low-resolution albedo map with low-frequency details; see \ref{sec:UV Albedo Parametric Model} for details.

In stage 3, a high-resolution albedo map with high-frequency details is generated. The detail albedo generator is driven by $\alpha_d$ to generate V-channel variations after fusing the rough UV albedo map V-channel. See \ref{sec:Detailed albedo generation} for details.

\begin{figure}[H]
\centering
\includegraphics[width=0.8\textwidth]{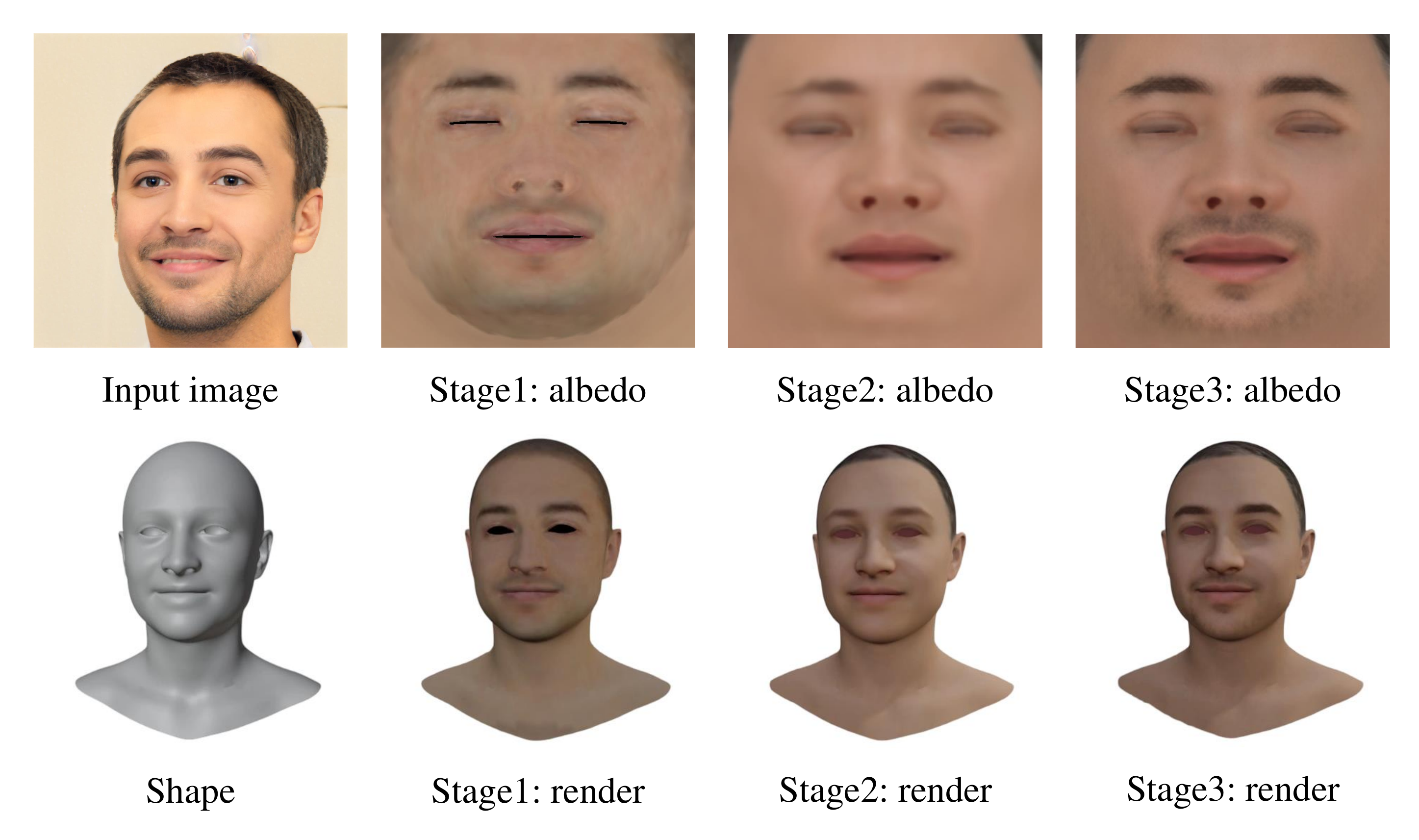}
\caption{Intermediate reconstruction results for each stage in our method. By comparing the rendered face with the input face, it can be seen that the final result is closer to the input face, especially the texture around the beard, eyebrows, and eye sockets.}\label{fig:stage_results}
\end{figure}

Finally, the rendered image is obtained by the differentiable renderer, which iteratively optimizes the coefficients of each stage by minimizing the loss function. Figure \ref{fig:stage_results} illustrates the intermediate results for each stage.

\subsection{UV Albedo Parametric Model}  \label{sec:UV Albedo Parametric Model}
Based on previous work, we follow 3DMMs to produce UVAPM from the public dataset FFHQ-UV which has UV layout consistency, uniform illumination, high resolution, and diversity. Since, directly unfolding the effective pixels in UV albedo maps into high dimensions according to a uniform rule will bring huge computational pressure, high memory usage, and long computation time. We split a UV map into $3$ channels, and then compute the PCA base and mean of each channel separately. An albedo map can be considered as a concatenation of $3$ independent channels:
$$
A_c = \sum_{i\in \{R,G,B\}} A^{(i)} = \sum_{i \in \{R, G, B\}} \left(\mathbf{\overline{A^{(i)}}}+ \alpha_{c}^{(i)} \mathbf{B_{alb}^{(i)}} \right),
$$
which $A_c\in\mathbb{R}^{3\times d\times d}$ and $d$ denote the UV coarse albedo map and its size, respectively. $R$, $G$, and $B$ denote the red, green, and blue channels, respectively. $A^{(i)} \in \mathbb{R}^{1\times d\times d}$ denotes $i$ channel of $A_c$. $\mathbf{\overline{A^{(i)}}} \in [0,255]$, $\alpha_{c}^{(i)}\in \mathbb{R}^{100}$ and $\mathbf{B_{alb}^{(i)}}$  representing the mean, coarse coefficients and PCA base of $i$ channel in the UV coordinate system, separately. 

\begin{figure}[H]
\centering
\includegraphics[width=0.5\textwidth]{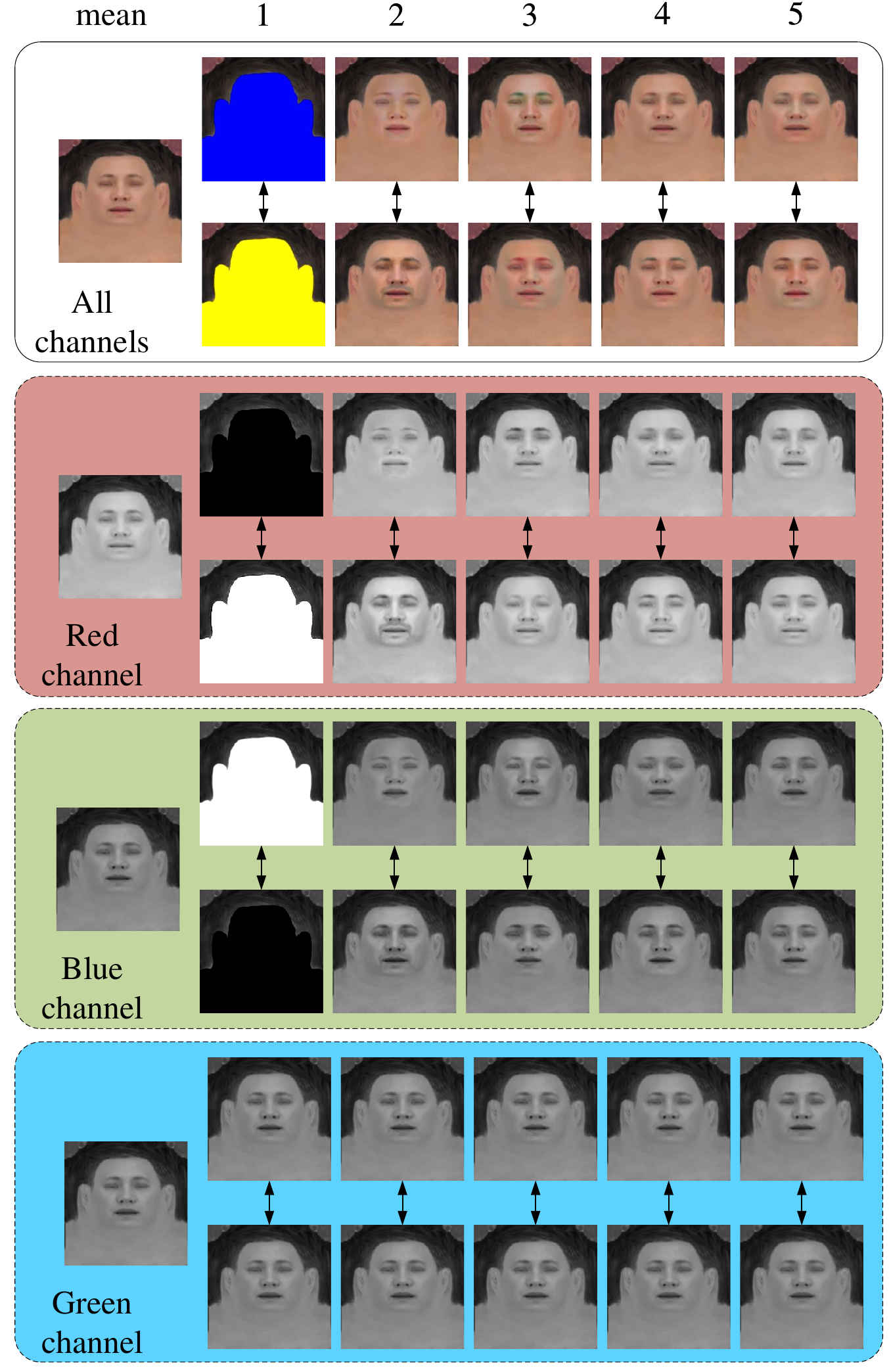}
\caption{Visualization of the UVAPM: the mean and the first five principal components, each of which is visualized as an addition and subtraction of the mean.}\label{fig:UVAPM}
\end{figure}

Figure \ref{fig:UVAPM} Visualization of the UVAPM components by showing the mean albedo map and the first five principal components of albedo variation. We can observe that the different principal components control the base skin color, and the main changing parts of the face (e.g., beard, eyebrows, eye sockets, etc.), respectively. Similarly, the figure further visualizes the first five principal components for each channel, each of which describes different attributes and features.

\subsection{Detailed albedo generation} \label{sec:Detailed albedo generation}

\begin{figure}[h]
\centering
\includegraphics[width=0.5\textwidth]{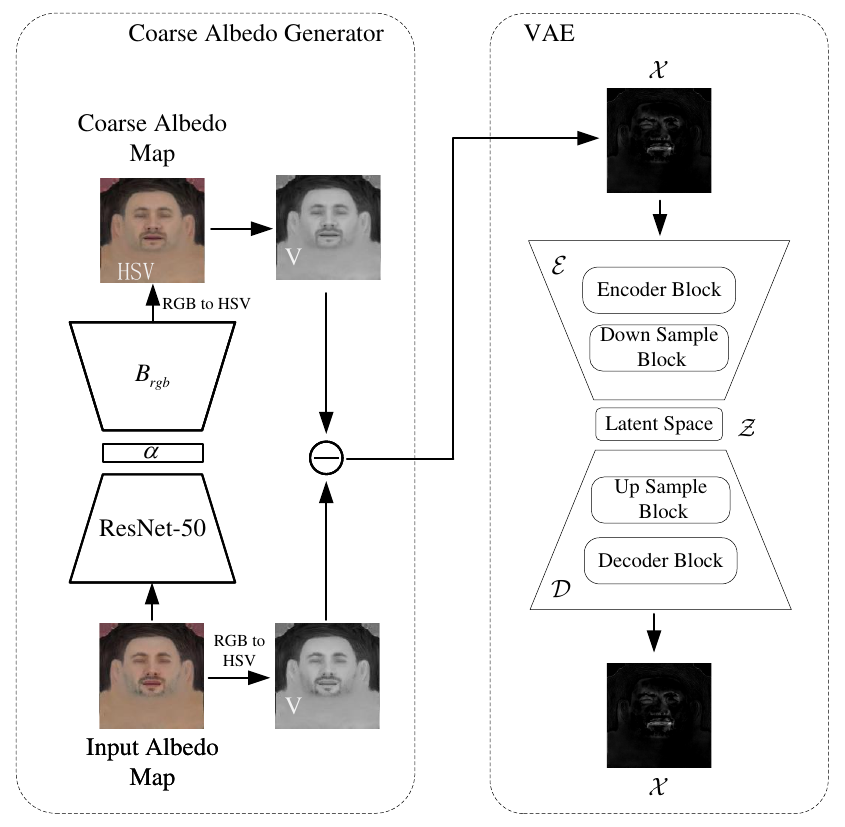}
\caption{The details of training detail generator}\label{fig:VAE_training}
\end{figure}  
HSV color space contains $3$ channels, Hue, Saturation, and Value, which can visually express the hue, vividness, and lightness of colors, and facilitate color comparisons, which is closer to the human experience of color perception. Since high-frequency detail features are reflected mainly in the V channel, we can change the V channel of the coarse albedo map $A_c$ to increase facial details, e.g. beard, forehead, cheeks, eye bags, etc.

We trained the detail generator on the detail dataset as shown in Figure \ref{fig:VAE_training}. To decouple the offsets of the V channels, we first trained a coarse albedo encoder using ResNet-50 and $\mathbf{B_{alb}}$ as decoder, forming an encoder-decoder training architecture. This architecture yields albedo maps with low-frequency information and facial skin tones in a high-resolution UV albedo dataset. Subsequently, we subtract the albedo maps generated by the trained coarse albedo generator from the original maps to obtain high-frequency detail maps. Finally, following \cite{LDM}, we train a VAE composed of an encoder $\mathcal{E}$ and a decoder $\mathcal{D}$ as a perceptual compression model, obtaining a low-dimensional latent space embedding $\mathcal{Z}=\mathcal{E}(x)\in\mathbb{R}^{32\times32\times4}$ that matches the UV texture data. Compared with the high-dimensional UV pixel space, this latent space makes full use of the low-dimensional representation of the UV space and is more suitable for training the likelihood-based generative model while effectively reducing memory consumption. Our albedo VAE can capture the details of UV albedo maps, making it possible to use it as a pre-train model in other 3D face reconstruction tasks.
We can get the details through:
$$
A_d = \mathcal{D}(\alpha_d),
$$
where $A_d$ denotes the details generated from $\mathcal{D}$.
The final albedo map can be expressed as:
$$
A = A_c^{(H,S)}\bigoplus (A_c^{(V)}+A_d) 
=A_c^{(H,S)}\bigoplus (A_c^{(V)}+\mathcal{D}(\alpha_d)), 
$$
Here, $A_c^{(i)}$ represents the value of channel $i\in\{H,S,V\}$ of $A_c$, which is obtained from the color space conversion function $\mathcal{T}_{hsv}$. $\bigoplus$ denotes channel concatenation. 

\subsection{Loss functions} \label{sec:loss_functions}
The overall loss function is defined as:
$$
\mathcal{L}
=\lambda_{pho}\mathcal{L}_{pho}
+\lambda_{lmk}\mathcal{L}_{lmk}
+\lambda_{id}\mathcal{L}_{id}
+\lambda_{mfc}\mathcal{L}_{mfc}
+\lambda_{reg}\mathcal{L}_{reg},
$$
where $\mathcal{L}_{pho}, \mathcal{L}_{lmk}, \mathcal{L}_{id}, \mathcal{L}_{mfc}, \mathcal{L}_{reg}$ are the photometric loss, landmark loss, identity loss, multi-view feature consistency loss and regularization item respectively with their corresponding weights $\lambda_{pho}, \lambda_{lmk}, \lambda_{id}, \lambda_{mfc}, \lambda_{reg}$. The different losses and the values of their corresponding weights are discussed in subsequent sections in detail. 

\subsubsection{Photometric loss}
The photometric loss function calculates the difference in color between the rendered image $I_r$ and the input image $I$ pixels. Since, different parts of the reconstruction are concerned differently, in this paper, various parts are weighted using the face mask approach and the face mask map can be obtained by the facial semantic segmentation method. 
$$
\mathcal{L}_{pho}=\frac{\sum\limits_{(i,j)\in I}G_{(i,j)}W_{(i,j)}\left\| {{I}_{(i,j)}}-{{I}_{r(i,j)}} \right\|_2}{\sum\limits_{(i,j)\in I}G_{(i,j)}},
$$
$W_{(i,j)}$ denotes the pixel facial mask weight located on $(i,j)$, which $G_{(i,j)}$ is $1$ indicates the pixel in the visible facial skin region and $0$ in elsewhere separately. Since the loss is computed only for the facial region, it improves the robustness of the model against common occlusions such as hair, clothes, sunglasses, etc. to a certain extent. We use face parsing\cite{face_parsing} to generate and select region-of-interest as facial masks, providing robustness to common occlusions by hair or other accessories.

\subsubsection{Landmark loss}
Face landmarks mainly contain the contours of facial parts (e.g., cheeks, nose, mouth, eyes, etc.), which can help the network learn the correct pose and contours. Landmark loss measures the difference between ground truth 2D facial landmarks and 2D landmarks projected from reconstructed 3D models of the landmarks by a camera model. The landmark loss is defined as:
$$
\mathcal{L}_{lmk}=\sum_{i=1}^{68}\omega_i (k_{gt}^i - k^i)^2,
$$
Here, We used 68 facial landmarks as constraints. The importance of landmarks varies across the face, and $\omega_i$ denotes the weight of the $i_{th}$ landmark. $k_{gt}^i\in\mathbb{R}^2$ and $k^i\in \mathbb{R}^2$ denote the coordinate of $i_{th}$ in $I$ and $I_r$, respectively. The pre-trained landmark detector\cite{landmark_detector} was used to detect $68$ keypoints during training.

\subsubsection{Identity loss}
Identity loss is widely used in 3D face reconstruction to enhance the identity similarity between the rendered image and the input image and to increase the fidelity of 3D face reconstruction. A pre-trained face recognition network \cite{vggface2_face_recognition} as $\mathcal{F}$  is used to extract feature embeddings from the original and rendered images. The loss is computed as the cosine similarity between these embeddings, which drives the rendered image to accurately capture the identity of the input image. The loss is defined as
$$
\mathcal{L}_{id}=1-\frac{\mathcal{F}(I)\mathcal{F}(I_r)}{\|\mathcal{F}(I)\|_2\cdot\|\mathcal{F}(I_r)\|_2},
$$

\subsubsection{Multi-view feature consistency loss}
Two images(i.e. $I^{i}$,$I^{j}$) taken from different viewpoints of the same subject at a given moment should regress with less difference in identity and expression coefficients. We follow the DECA strategy\cite{DECA}, randomly swapping the identity and expression coefficients regressed from different viewpoints and keeping the other coefficients constant. The multi-view feature consistency loss is defined as:
$$
\mathcal{L}_{mfc}=\mathcal{L}(I^{i}, \mathcal{R}(S(\beta^j, \xi^j, c^i), T(\alpha^i, \gamma^i, N_{uv}^i ))),
$$

\subsubsection{VAE loss}
Training of the variational auto-encoder by optimizing the loss term:
$$\mathcal{L}_{VAE}=\lambda_{kl}\mathcal{L}_{KL}+\lambda_{gan}\mathcal{L}_{GAN}+\lambda_{lpips}\mathcal{L}_{LPIPS},
$$
Here, $\mathcal{L}_{KL}$ denotes the KL divergence loss\cite{VAE}, $\mathcal{L}_{GAN}$ denotes the GAN adversarial loss\cite{GAN_loss}, $\mathcal{L}_{LPIPS}$ denotes the LPIPS perceptual loss\cite{LPIPS}. The loss weights $\lambda_{kl}$, $\lambda_{gan}$, and $\lambda_{lpips}$ control the relative contributions of the different loss terms during optimization.

\subsubsection{Regularization}
$\mathcal{L}_{reg}$ regularizes coefficients of each submodule, by minimizing the L2 loss of $\beta, \xi, \gamma, \alpha$

\section{Experiments}
\subsection{Implementation details}
We use the PyTorch deep learning framework and take advantage of the differentiable renderer\cite{differentiable_renderer} for rendering. We train our model for 50 epochs on NVIDIA 4090D GPUs with a mini-batch of 32. we use Adam\cite{diederik2014adam} as optimizer with an initial learning rate of $1e-3$. The input image is cropped and aligned by\cite{chen2016supervised} and resized into $224\times224$. 
\subsection{Evaluation metrics}
We utilize the Learned Perceptual Image Patch Similarity (LPIPS) \cite{LPIPS} as a metric to assess reconstruction accuracy. LPIPS is a deep learning-based measure that effectively captures perceptual differences between images, thereby providing a more precise evaluation of reconstruction quality. Additionally, to quantify the visual quality of the generated images, we employ the Fréchet Inception Distance (FID) \cite{FID}. FID is a widely used metric in the evaluation of Generative Adversarial Networks (GANs), which compares the distribution of features extracted from real and generated images in a feature space, effectively reflecting the quality of visual output.

Furthermore, to evaluate the model's ability to retain identity information, we adopt the cosine similarity (CSIM) \cite{CSIM1, CSIM2} of facial identity embeddings. Cosine similarity is a commonly used metric that measures the cosine of the angle between different facial embedding vectors, effectively gauging the similarity of facial features and reflecting the model's performance in preserving identity information. 

Two evaluation metrics are used, the Peak Signal-to-Noise Ratio (PSNR) and the Structure Similarity Index Measure (SSIM). These are computed by comparing the predicted UV texture with the ground truth.

\subsection{Qualitative comparison}

\begin{figure}[H]
\centering
\includegraphics[width=0.8\textwidth]{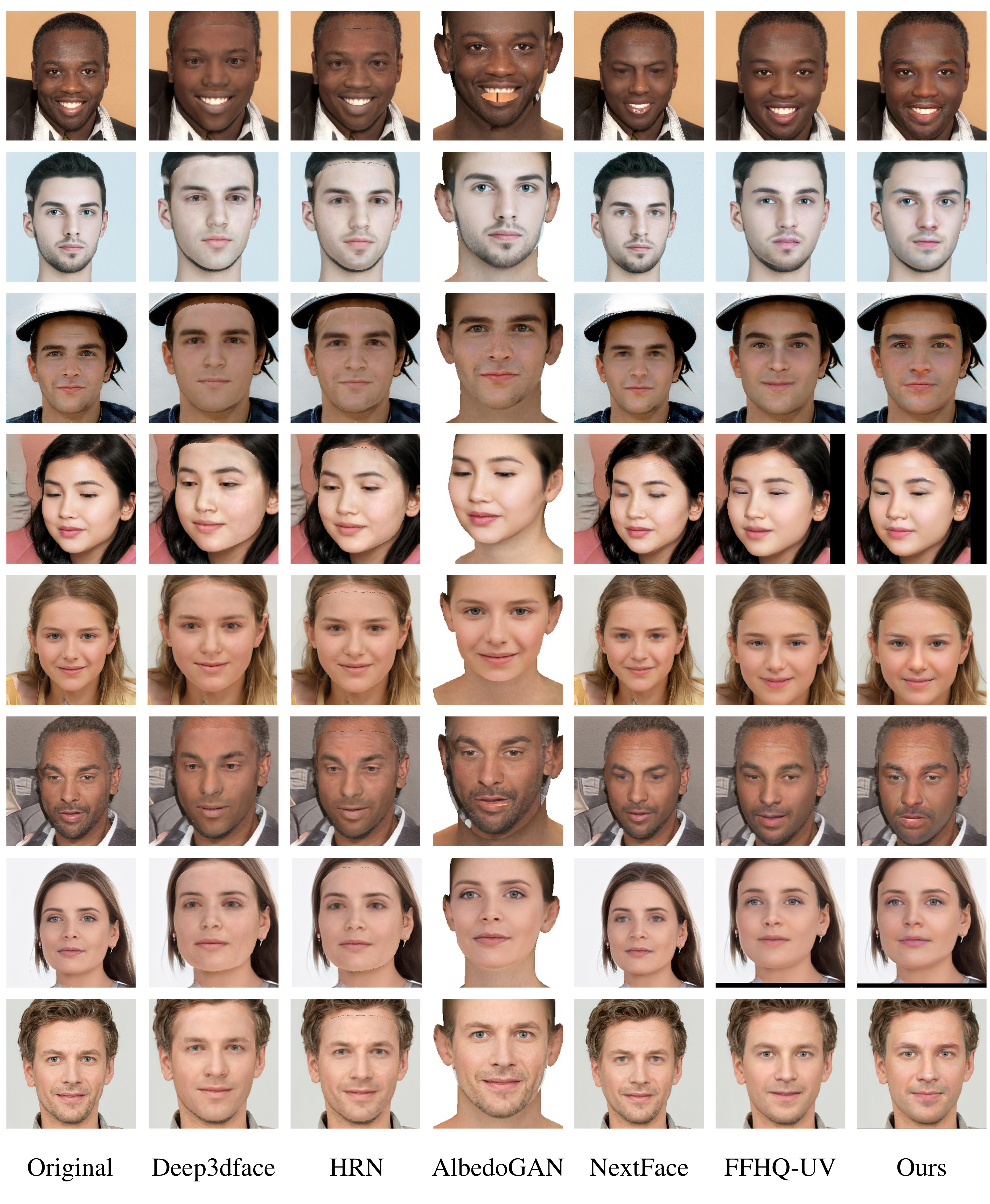}
\caption{Comparison of rendered images using different methods.}\label{fig:Comparison of rendered images}
\end{figure}

Our approach is qualitatively compared with state-of-the-art (SOTA) face texture generation methods, examples of which are shown in Figure \ref{fig:Comparison of rendered images}. Examples are randomly selected from the AI-10K dataset generated using StyleGAN3\cite{StyleGAN3}, which include different genders, ages, skin tones, poses, and expressions, a total of $10,000$.

It can be seen that despite careful optimization of Deep3D, the final texture quality is still limited by the ability of the linear base representation, resulting in blurred textures that ignore high-frequency features of the face.
The GAN-based method FFHQ-UV uses a multi-stage parameter optimization strategy, where the predicted shape and the generated albedo maps are optimized simultaneously in the 3rd stage, and although high-resolution UV albedo maps can be generated, some facial parts are overfitted, such as the corners of the mouth in the 1st and last rows of Figure \ref{fig:Comparison of rendered images}.
HRN\cite{HRN} is a multi-stage coarse-to-fine method that jointly derives and optimizes the geometry and texture, and although it looks similar to the input image on the whole, the generated texture is smooth, performs poorly in the highlights, and fails to capture complex facial details such as the male whiskers in the 2nd and 6th rows of Figure \ref{fig:Comparison of rendered images}.
Since the test code for the GAN-based method AlbedoGAN\cite{AlbedoGAN} is not yet fully publicly available, we can only demonstrate the texture reconstruction effect of this method through its provided technique for extracting texture mapping from the original image. As seen in the 1st and 6th rows of Figure \ref{fig:Comparison of rendered images}, the accuracy of the face alignment directly affects the texture rendering.
Overall, the image rendered by the iterative fitting-based method NextFace\cite{NextFace} closely resembles the input image. However, it exhibits instability in the darker or occluded areas of the face, such as the eye socket region shown in the 1st row of Figure \ref{fig:Comparison of rendered images}. Additionally, the expressive power of the low-resolution albedo map limits its ability to depict high-frequency details accurately. As illustrated in Figure \ref{fig:Comparison of rendered images}, our method achieves more precise and high-fidelity results in both shape and texture in these scenarios.

\begin{figure}[H]
\centering
\includegraphics[width=0.8\textwidth]{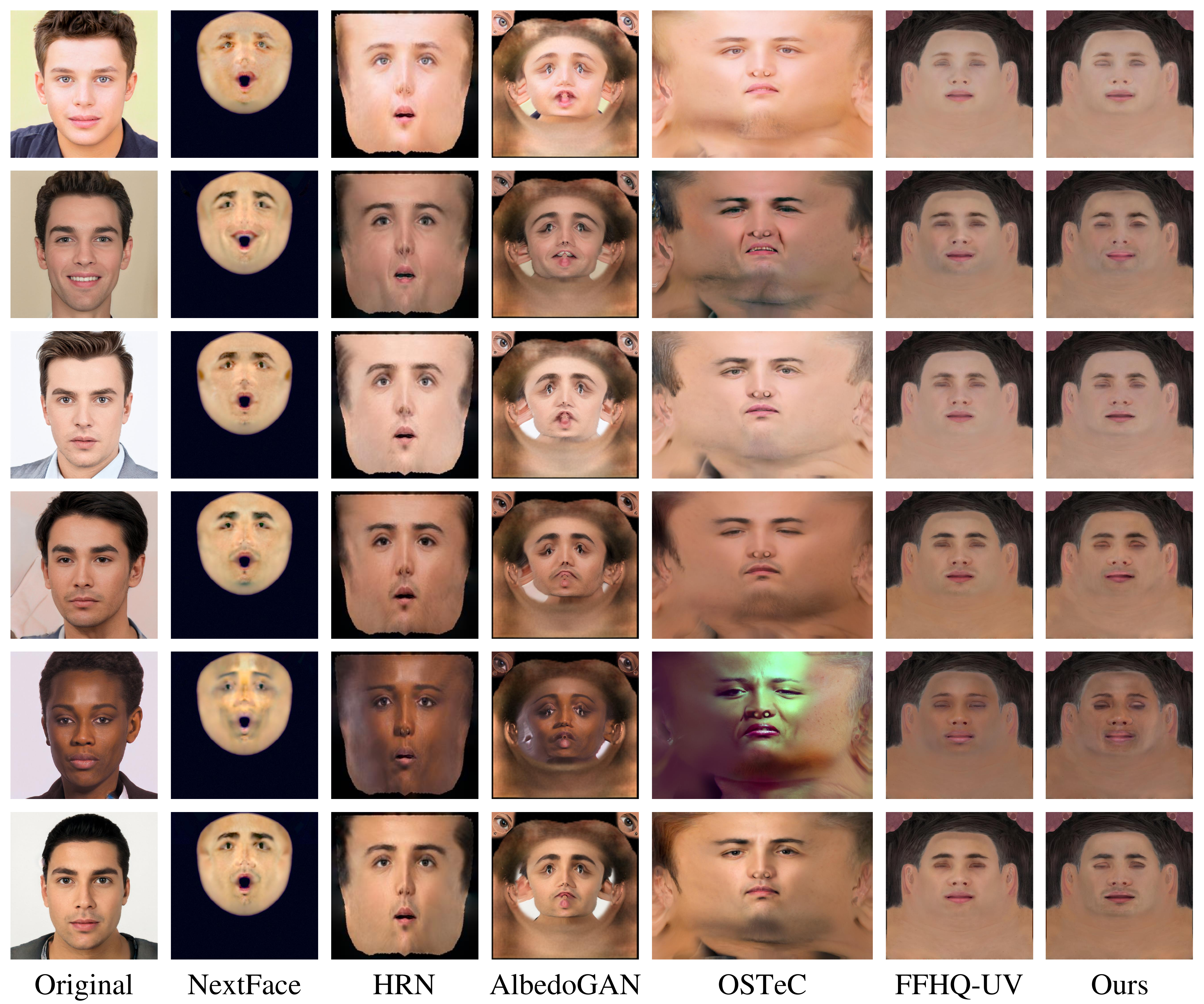}
\caption{Visual comparisons of texture generation using NextFace, HRN, AlbedoGAN, OSTeC, FFHQ-UV, and our method.}\label{fig:Visual comparisons of texture generation}
\end{figure}

We conducted a comparative analysis of the UV texture maps generated by our method and state-of-the-art (SOTA) methods. For this evaluation, we utilized AI-10K as test samples, ensuring that all input images maintained a consistent size of $224\times224$ pixels for fairness. The output texture size is influenced by the underlying topology and the UV texture mapping formulas, allowing us to identify structural differences in the textures produced by various methods.
As illustrated in Figure \ref{fig:Visual comparisons of texture generation}, the texture-completion method, OSTec, rotates the input image in 3D and fills in the invisible regions by reconstructing the rotated image based on the visible parts in the 2D face generator. However, its ability to reconstruct darker faces is somewhat limited, as evidenced in the 5th row of Figure \ref{fig:Visual comparisons of texture generation}, where the refined UV texture map exhibits distorted colors due to overfitting lighting conditions.
The extraction-based method, AlbedoGAN, which aims to extract the original texture, results in a less effective final UV texture map, particularly in the invisible cheek area. This limitation arises from the accuracy of face alignment during the extraction process.
Fitting-based methods NextFace and HRN, produce textures that are smaller in size, exhibit smooth surfaces, and lack high-frequency detail. In contrast, FFHQ-UV generates high-frequency details but also displays artifacts, particularly in the forehead region. Our method demonstrates superior performance in handling highlights, cheeks, and invisible areas, effectively addressing the limitations observed in other approaches.

\subsection{Quantitative comparison}

We present a quantitative comparison with state-of-the-art (SOTA) methods in Table \ref{tab1}. NextFace relies on iterative fitting and is susceptible to overfitting in the presence of occlusions; therefore, we excluded this approach from our quantitative experiments. To comprehensively evaluate the performance of the methods, we constructed an evaluation dataset comprising 100 images selected from the FaceScape and AFLW\cite{AFLW2000_dataset} datasets. This dataset is evenly divided, with half consisting of controlled face images and the other half comprising in-the-wild images that exhibit a uniform distribution of facial occlusion sites, poses, lighting conditions, expressions, highlights, genders, and age attributes, none of which appear in the training dataset. To mitigate discrepancies between input and output image sizes, we normalized all images to a resolution of $224\times224$ pixels.
As illustrated in Table \ref{tab1}, our method demonstrates superior performance compared to Deep3D and HRN in terms of LPIPS, FID, and CSIM metrics, highlighting its exceptional capabilities in reconstruction, generation, and identity preservation. The AFLW dataset encompasses a substantial number of facial images captured in complex scenes, necessitating a texture generator with enhanced capabilities. Although our CSIM score is slightly lower than FFHQ-UV in this dataset, our method exhibits advantages in albedo reconstruction and rendering quality. OStec proposes an unsupervised single-shot 3D face texture completion method. The method obtains a complete texture by filling the invisible region with the rotated image, however, the final rendered face image is frontal and the background is not the input background, thus it performs poorly in all metrics. Additionally, we included low-dimensional UV texture bases in our comparison, revealing that these bases perform worse than their high-dimensional counterparts across all evaluation metrics.

\begin{table}[h]
\centering
\caption{Quantitative comparison of rendered face quality}\label{tab1}%
\begin{tabular}{@{}cccccccc@{}}
\toprule
\multirow{2}{*}{Method} & \multirow{2}{*}{FID$\downarrow$} & \multicolumn{3}{c}{LPIPS$\downarrow$} & \multicolumn{3}{c}{CSIM$\downarrow$} \\
\cmidrule{3-5}\cmidrule{6-8}
 &  & avg. & med. & std. & avg. & med. & std. \\
\midrule
OSTeC\cite{Ostec}    & 221.1215 & 43.6577e-2 & 43.3400e-2 & 29.0948e-4 & 235.1610e-3 & 213.3737e-3 & 669.8990e-5 \\
HRN\cite{HRN}      & 32.3890  & 7.9029e-2  & 7.2895e-2  & 5.5028e-4  & 6.1732e-3   & 4.6419e-3   & 4.5767e-5   \\
Deep3D\cite{Deep3dface}   & 28.8910  & 7.5559e-2  & 6.8660e-2  & 5.9677e-4  & 8.1783e-3   & 5.9777e-3   & 6.7479e-5   \\
FFHQ-UV\cite{FFHQ-UV}  & 25.5443  & 4.8682e-2  & \textbf{4.0845e-2}  & 5.2443e-4  & \textbf{3.1825e-3}   & \textbf{2.2542e-3}   & \textbf{0.9525e-5}   \\
\midrule
UVAPM-128 & 22.0238  & 4.9318e-2  & 4.2650e-2  & 4.7216e-4  & 3.5313e-3   & 2.8110e-3   & 1.2905e-5   \\
UVAPM-256 & \textbf{21.8297} & \textbf{4.8526e-2} & 4.1083e-2 & \textbf{4.9122e-4} & 3.6087e-3 & 2.7477e-3 & 1.4376e-5 \\
\botrule
\end{tabular}
\end{table}

\subsection{Ablation study}

We emphasize the importance of high linear texture base dimensions and detailed texture generation by comparing the metrics results for linear UV texture bases of different accuracies and the corresponding addition of detail. For a valid comparison, we avoided the rendering process to prevent interference and trained the encoders with the same hyperparameters. Specifically, we compare the difference between the output and the input under the same test dataset of 500 randomly selected images from FFHQ-UV, which were not present during training and contain different genders, ages, skin tones, etc. Multiple evaluation metrics are used for the overall evaluation, and these metrics contain both pixel level and perceptual level. As can be seen from the different sets of results in Table \ref{tab3}, the addition of high-frequency detailed textures to coarse textures resulted in considerable improvement in all the results. The increase in the linear texture base dimension allows for a richer representation of the tones on the face.

\begin{figure}[H]
\centering
\includegraphics[width=1.0\textwidth]{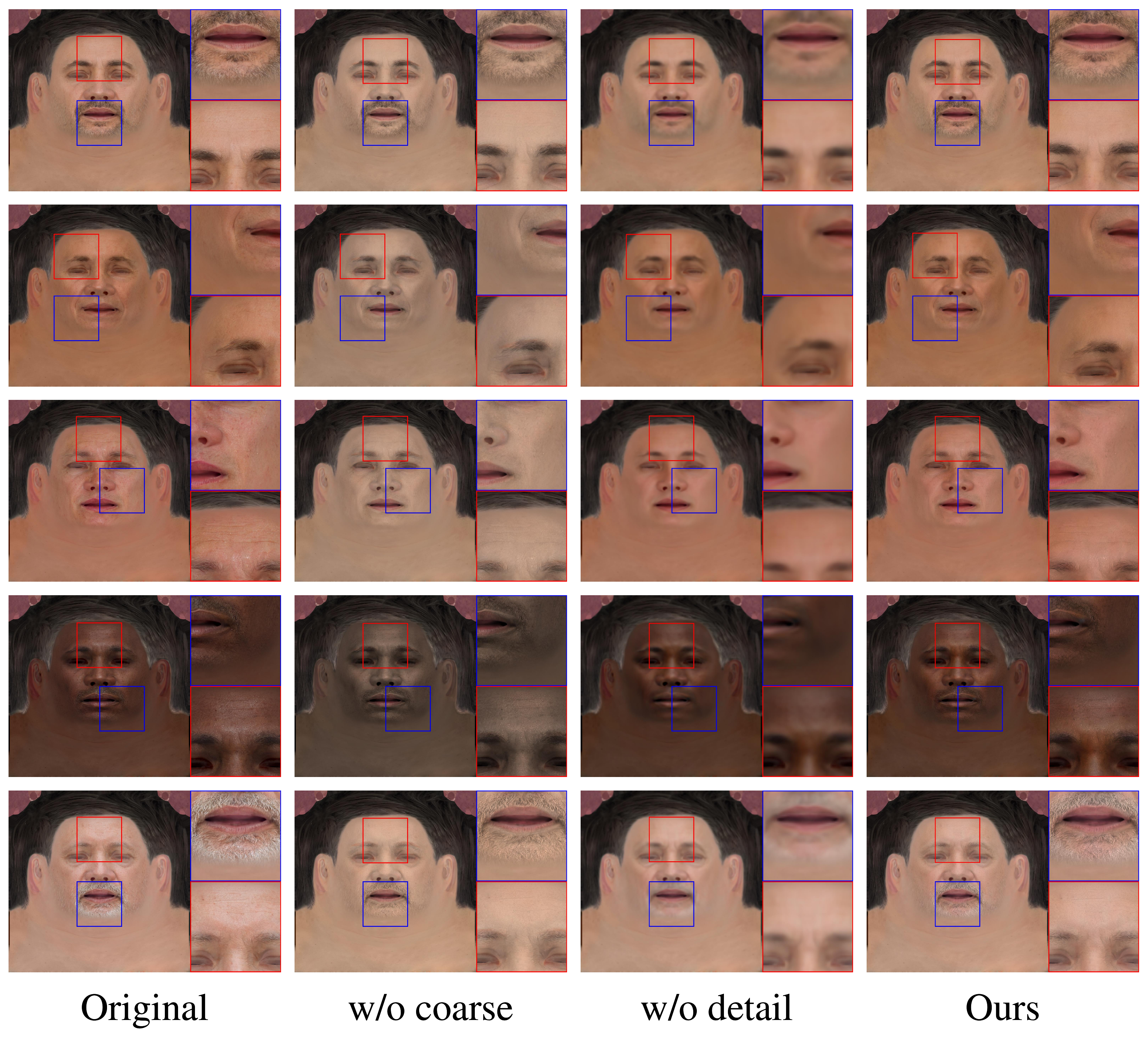}
\caption{The visual examples of our ablation analysis.}\label{fig:visual examples of ablation}
\end{figure}

To demonstrate the effectiveness of the linear texture base and the detail texture generator, we visualized the difference between the textures regressed without coarse and without detail. Specifically, we selected a few UV texture images with complex facial textures that contain rich high-frequency details such as wrinkles, whiskers, spots, highlights, etc. To conveniently demonstrate the effect of detail texture, we migrate the regressed-out details to the template face.
The visualization shows that without the addition of high-frequency detail texture, the coarse texture maps regressed from the UV texture base exhibit blurring, especially at the whiskers and the wrinkles at the corners of the mouth, as shown in the 1st, 2nd, and last rows in Figure \ref{fig:visual examples of ablation}.
When the facial tones regressed from UVAPM are removed, it is difficult to ensure identity, even with the addition of high-frequency details. As shown in Figure \ref{fig:visual examples of ablation}, the last row of the beard contains only black whiskers and does not contain white whiskers.

\begin{table}[h]
\centering
\caption{Ablation studies}\label{tab3}
\begin{tabular}{@{}c|c|cccc@{}}
\toprule
Coarse              & Detail        & FID$\downarrow$       & MSE$\downarrow$       & PSNR$\uparrow$       & SSIM$\uparrow$                  \\
\midrule
$64\times64$        & $\times$      & 248.83     & 50.28      & 31.14      & 0.88                     \\
$64\times64$        & \checkmark    & 27.18      & 9.36       & 38.63      & 0.97                     \\
\midrule
$128\times128$      & $\times$      & 133.78     & 27.97      & 33.71      & 0.92                     \\
$128\times128$      & \checkmark    & 23.02      & 21.17      & 34.97      & 0.97                     \\
\midrule
$256\times256$      & $\times$      & 76.86      & 14.35      & 36.65      & 0.95                     \\
$256\times256$      & \checkmark    & 20.37      & 7.82       & 39.47      & 0.98                     \\
\botrule
\end{tabular}
\end{table}

\section{Conclusion and Future Work}
\subsection{Conclusion}

We have introduced an end-to-end coarse-to-fine albedo generation approach that generates textures that can be directly used in HiFi3Ds-based 3D face reconstruction methods. In particular, UVAPM is produced from the high-quality dataset FFHQ-UV, which can be driven by low-dimensional coefficients to obtain an albedo map with only low-frequency texture information. In order to reconstruct an albedo map with high-frequency texture information, the detail generator is proposed for generating details. the reconstructed albedo map can be used for rendering in different ambient light. The code and trained albedo generator will be publicly available on GitHub.

\subsection{Limitations and future work}
The proposed UVAPM was currently attempted to be built only at 3 lower resolutions, and despite our strategy of calculating by subchannels, we still cannot effectively solve the law that the computational pressure increases exponentially with the elevation of dimensionality. In addition, we believe that simply boosting the PCA base richness still cannot escape the fitting ability of linear models. Therefore, this work used nonlinear generative capabilities to compensate for the shortcomings of linear models. In the shape reconstruction task, only linear shape basis and expression basis are used to regress the face model, and no further displacement maps are included, which is not the focus of this work and has little impact, but in the future, we will explore more accurate geometric models. Finally, in this paper, only coefficient fitting is used to generate albedo maps, and the use of self-supervised learning methods will be explored in the future.

\subsection{Acknowledgements}
This work has been partly supported by Natural Science Research in Colleges and Universities of Anhui Province of China under Grant Nos.KJ2020A0362.

\bibliography{bibliography}
\end{document}